\documentclass[twoside, 11pt]{article}

\usepackage{jmlr2e}
\usepackage{geometry}
\usepackage[utf8]{inputenc} 
\usepackage[T1]{fontenc}    
\usepackage{hyperref}       
\usepackage{url}            
\usepackage{booktabs}       
\usepackage{amsfonts}       
\usepackage{nicefrac}       
\usepackage{microtype}      
\usepackage{amsmath}
\usepackage{bbm}
\usepackage{dsfont}
\usepackage{color}
\usepackage[ruled,vlined,linesnumbered]{algorithm2e}
\usepackage{algcompatible}
\usepackage{xcolor}

\global\long\def\argmin{\operatornamewithlimits{argmin}}

\newcommand{\xb}{\underline{x}}

\newcommand{\E}{\mathbbm{E}}

\global\long\def\argmin{\operatornamewithlimits{argmin}}

\ShortHeadings{}{}
\firstpageno{1}

\begin{document}

\title{Learning Continuous Exponential Families Beyond Gaussian}

\author{\name Christopher X. Ren \email cren@lanl.gov \\
      \addr Intelligence and Space Research Division\\
      Los Alamos National Laboratory\\
      Los Alamos, NM 87545, USA
      \AND
      \name Sidhant Misra \email sidhant@lanl.gov \\
      \addr Theoretical Division\\
      Los Alamos National Laboratory\\
      Los Alamos, NM 87545, USA
      \AND
      \name Marc Vuffray \email vuffray@lanl.gov \\
      \addr Theoretical Division\\
      Los Alamos National Laboratory\\
      Los Alamos, NM 87545, USA
      \AND
      \name Andrey Y.\ Lokhov \email lokhov@lanl.gov \\
      \addr Theoretical Division\\
      Los Alamos National Laboratory\\
      Los Alamos, NM 87545, USA}

\editor{}

\maketitle

\begin{abstract}
We address the problem of learning of continuous exponential family distributions with unbounded support. While a lot of progress has been made on learning of Gaussian graphical models, we still lack scalable algorithms for reconstructing general continuous exponential families modeling higher-order moments of the data beyond the mean and the covariance. Here, we introduce a computationally efficient method for learning continuous graphical models based on the Interaction Screening approach. Through a series of numerical experiments, we show that our estimator maintains similar requirements in terms of accuracy and sample complexity scalings compared to alternative approaches such as maximization of conditional likelihood, while considerably improving upon the algorithm's run-time.
\end{abstract}


\section{Introduction}

An exponential family is a general parametric set of probability distributions that are capable of capturing any positive probability measure \cite{koopman1936distributions}. Graphical models that aim to model conditional dependencies of high-dimensional probability distributions with discrete or continuous variables are commonly expressed in the exponential family form \cite{wainwright2008graphical}. Exponential family distributions are typically characterized by the energy function which specifies the sufficient statistic of the distribution, the associated parameters, and the graph structure of the underlying graphical model:
\begin{equation}
    \mu(\xb) = \frac{1}{Z} \exp \left[ E(\xb) \right],
    \label{eq:General_Distribution}
\end{equation}
where $\xb \in \mathbb{R}^p$ denotes a collection of $p$ random variables, $E(\xb)$ is the energy function, and $Z$ is the normalization factor known as the partition function. The energy function takes the general form
\begin{equation}
    E(\xb) = - \sum_{k \in \mathcal{K}} \theta_k f_k(\xb_k),
    \label{eq:General_Energy_Function}
\end{equation}
where $\{f_k, \ k \in \cal{K} \}$, for some set $\cal{K}$, is a collection of basis functions each acting upon subsets of variables $\xb_k$. These basis functions are often referred to as the sufficient statistic of the model. The parameters $\theta_k \in \mathbb{R}$ specify a particular model within this family.

Perhaps the most famous continuous and unbounded exponential family is the multivariate normal distribution with mean $\mu$ and covariance $\Sigma$. When defined on a $p$-dimensional graph $G=(V,E)$ with $|V|=p$, the energy function of a Gaussian Graphical Model (GGM) can be expressed as a quadratic function
\begin{align}
    \nonumber
    E(\xb) = & -\frac{1}{2} \sum_{i \in V} \theta_{ii}(x_i - \mu_i)^2
    \\
    &- \sum_{(i,j) \in E} \theta_{ij}(x_i - \mu_i) (x_j - \mu_j),
    \label{eq:Gaussian_energy_function}
\end{align}
where $\mu_i$ denotes the mean of the variable $x_i$, and $\theta_{ij}$ are elements of the \emph{precision matrix} $\Theta$ whose support is determined by the sparsity pattern of the graph $G$. GGMs have a special property that $\Theta$ is equal to the inverse of the covariance matrix $\Sigma$, meaning that $(\Sigma^{-1})_{ij} = 0$ for all $(i,j) \notin E$. GGMs are frequently used in applications where the random variables arise from sums of independent contributions and hence are well approximated by a multivariate Normal distribution due to the central limit theorem.  However, numerous applications in natural sciences and artificial systems generate data with statistics that strongly deviate from Gaussian distributions, where higher-order moments are important \cite{banfield1993model}. Examples of such non-Gaussian statistics include turbulent flows \cite{li2005origin}, scattered waves \cite{jakeman1988non}, cosmic microwave background temperature \cite{kogut1996tests}, diffusion of biomacromolecules \cite{ghosh2016anomalous}, gene expression and bioinformatics \cite{ji2005applications}, atmospheric data \cite{perron2013climatology} and cloud data \cite{sengupta2016predictive} available through remote-sensing imagery, modeling of color image pixels in the RGB space \cite{bouguila2004unsupervised}, and non-negative matrix factorization in recorded music \cite{10.5555/3104322.3104379}. Motivated by many applications which aim to explicitly model higher-order moments present in the data, it is natural to consider a generalization of GGMs consisting of adding higher-order sufficient statistics to the energy function. An example of such a generalization is a graphical model that models moments up to four:
\begin{align}
    \nonumber
    E(\xb) = - &\sum_{i} \theta_{i} x_i - \sum_{ij}\theta_{ij} x_i x_j \\
    - &\sum_{ijl}\theta_{ijl} x_i x_j x_l
    - \sum_{ijlm}\theta_{ijkl} x_i x_j x_l x_m.
    \label{eq:General_Energy_Function_Example}
\end{align}
More generally, we can write the energy function with sufficient statistics $f_k$ consisting of monomials acting upon subsets of variables $\xb_k$. 
In the example \eqref{eq:General_Energy_Function_Example} above, $\cal{K}$ is the set of multi-indices of size at most four and if $k=ijl$ we would have $f_k(\xb_k) = x_i x_j x_l$.
In this representation, the sparsity pattern of the array $\Theta = \{\theta_k, \ k \in \cal{K} \}$ provides interpretable structural information on the conditional dependencies between variables through the separation property of Markov Random Fields (MRFs). Given the rich structural information contained in the array of parameters, it is of interest to develop algorithms for reconstructing $\Theta$ and the underlying hypergraph from data. We consider the following learning problem.
\begin{definition}[Learning Problem]
Given $n$ i.i.d. samples $\{\xb^{(k)}\}_{k=1,\ldots,n}$ drawn from an unknown probability measure \eqref{eq:General_Distribution} with the general energy function \eqref{eq:General_Energy_Function} and monomial basis functions, estimate the parameter array $\Theta$.
\end{definition}
If $\Theta$ is learned to a sufficient accuracy, a thresholded version of $\Theta$ will uncover the underlying hypergraph, and hence the structure of the conditional dependencies through the respective sparsity pattern. In the remainder of the paper, we assume that $\Theta$ is such that the distribution in \eqref{eq:General_Distribution} exists, i.e. $\exp \left[ E(\xb) \right]$ is integrable. 

Many efficient algorithms of different types are known for learning Gaussian graphical models \cite{meinshausen2006high, tibshirani1996regression, yuan2007model, BanerjeeGhaoui2008, ravikumar2008model, anandkumar2012high, cai2011constrained, cai2012estimating}, including polynomial-time algorithms achieving information-theoretic bounds in terms of sampling complexity for general GGMs \cite{misra2020information} and important sub-classes of GGMs \cite{kelner2019learning}. In comparison, scalable  methods for learning general continous exponential family distributions beyond Gaussian are still lacking. Unlike the Gaussian setting, the maximum likelihood estimator is computationally intractable for general continuous graphical models: the computation of the normalization factor
\begin{equation}
    Z = \int_{-\infty}^{+\infty} \ldots \int_{-\infty}^{+\infty} \left( \prod_{i=1}^{N} dx_i \right) \exp \left[ E(\xb) \right]
\end{equation}
results in a hard high-dimensional integration problem \cite{koutis2003hardness}.

A popular approach to the learning of high-dimensional distributions for which the maximum likelihood is intractable consists of using a Pseudo-Likelihood (PL) method that consistently recovers the model in polynomial time \cite{besag1975statistical}. The PL approach is based on conditional likelihood maximization, and has been previously used for Gaussian distributions \cite{meinshausen2006high}, discrete graphical models \cite{ravikumar2010high, Klivans2017, lokhov2018science}, continuous generalized linear models \cite{yang2015graphical, yang2018semiparametric}, MRFs with pairwise potentials \cite{tansey2015vector, yuan2016learning, suggala2017expxorcist}, and mixed graphical models \cite{lee2015learning}. However, as we show in this work, the application of Pseudo-Likelihood to the challenging case of continuous distributions with unbounded support is hindered by the necessity of numerically computing local normalization factors, rendering this estimation method impractical even for a small number of random variables.

In practice, non-Gaussian statistics are often studied using other specific families of parametrized distributions, such as Beta, Dirichlet, Gamma, and Poisson distributions \cite{ma2011non}, or various mixture models \cite{titterington1985statistical}, for which tailored learning methods have been developed. Existing related works in the estimation of multivariate distributions that generalize Gaussian include semi-parametric methods, such as non-paranormal \cite{liu2009nonparanormal, lafferty2012sparse}, Gaussian copula models for a certain choice of copula functions \cite{dobra2011copula, liu2012high}, or rank-based estimators \cite{xue2012regularized} which use transformations that ``Gaussianize'' the data, and then fit GGMs to estimate model structure. Other approaches include score matching for reproducing kernel Hilbert space parametrizaton \cite{sun2015learning}, or a transport maps based approach shown in \cite{morrison2017beyond} which uses parametrization with Hermite polynomials.

In this work, we propose a general computationally efficient estimator for learning general continuous graphical models of the type \eqref{eq:General_Energy_Function}. Our method is based on a suitable modification and generalization of the \emph{Interaction Screening} method first introduced in \cite{Vuffray2016nips} for pairwise graphical models over binary variables, and recently generalized to discrete graphical models with arbitrary degree of interactions and alphabet in \cite{vuffray2019efficient} with improved sample-complexity on all existing methods. Unlike Pseudo-Likelihood, the Interaction Screening estimator does not include a normalization factor, and can be generalized to the case of noisy or corrupted data \cite{goel2019learning}. Building on the analysis of \cite{vuffray2019efficient}, recent works \cite{shah2020learning, shah2021computationally} applied the Interaction Screening technique to the setting of continuous graphical models, however restricted to distributions with pairwise interactions, and more importantly to distributions with bounded support, and showed an exponential dependence of sample complexity on the domain bound. Indeed, as we explain later in this paper, the Interaction Screening estimator as initially introduced in \cite{vuffray2019efficient} does not directly apply to continuous distributions with unbounded support.

The main goal of this paper is to introduce a practical estimator that overcomes all computational intractability problems of Pseudo-Likelihood and original Interaction Screening estimators for non-Gaussian distributions. We design a special regularization of the Interaction Screening estimator that ensures that the estimator is well-behaved in the setting of general continuous exponential families with non-pairwise energy functions and unbounded support.
We empirically show that the new \emph{Interaction Screening Objective for Distributions with Unbounded Support (ISODUS)} and the respective regularized estimator maintain similar scaling requirements in terms of accuracy and sample complexity compared to Pseudo-Likelihood, while avoiding the numerical evaluation of a large number (order $pn$) of integrals at every step of the learning procedure, hence significantly improving upon PL's run-time. Importantly, the new estimator demonstrates consistency properties similar to the state-of-the-art algorithms for learning exponential family distributions with bounded support, recovering the structure of any model without unnecessary assumptions such as incoherence or irrepresentability.

\section{Algorithms}

In this section, we start by presenting the benchmark algorithm based on the Pseudo-Likelihood approach \cite{besag1975statistical} that we will use as a baseline in our experiments, and discuss challenges associated with the optimization for non-Gaussian graphical models. Then, we introduce our new ISODUS estimator based on the Interaction Screening principle \cite{vuffray2019efficient}.

\subsection{Pseudo-Likelihood}
To introduce the Pseudo-Likelihood estimator, we first define the conditional distribution for a single variable $x_i$. For $i \in \{1,\ldots,p\}$, let $\mathcal{K}_{i} \subseteq \mathcal{K}$ denote the set of monomials acting upon subsets $\xb_k$ that contain the variable $x_i$.
Let $E_i(\xb)$ denote the terms in $E(\xb)$ that participate in $\mathcal{K}_{i}$:
\begin{equation}
    E_i(\xb) = - \sum_{k \in \mathcal{K}_i} \theta_k f_k(\xb_k).
    \label{eq:Local_energy_function}
\end{equation}
Then, the conditional distribution of $x_i$ reads
\begin{equation}
    P(x_i \mid \xb_{\backslash i}) = \frac{1}{Z_i} \exp \left[ E_i(\xb) \right],
    \label{eq:Conditional_distribution}
\end{equation}
where $\xb_{\backslash i}$ denotes the set of variables $\{x_i\}_{i \in V}$ with $x_i$ removed from the set. For the example introduced in Eq.~\eqref{eq:General_Energy_Function_Example}, the conditional distribution is
\begin{align*}
    P(x_i \mid & \xb_{\backslash i}) = \frac{1}{Z_i} \exp \Big(- \theta_{i} x_i - \sum_{j}\theta_{ij} x_i x_j\\
    &-\sum_{jl}\theta_{ijl} x_i x_j x_l - \sum_{jlm}\theta_{ijlm} x_i x_j x_l x_m \Big).
    \label{eq:Conditional_distribution_example}
\end{align*}
The normalization factor, which we refer to as the local partition function, cannot be explicitly computed in general, and is given by a one-dimensional integral:
\begin{equation}
    Z_i = Z_i(\xb_{\backslash i}) = \int_{-\infty}^{+\infty} dx_i \exp \left[ E_i(x_i \Vert \xb_{\backslash i}) \right],
    \label{eq:Integral_Normalization_Conditional}
\end{equation}
where notation $E_i(x_i \Vert  \xb_{\backslash i})$ implies that variables $\xb_{\backslash i}$ are fixed in this expression. Notice that the value of the normalization constant $Z_i$ depends on the model parameters $\Theta_i$, where $\Theta_i$ denote components of the array $\Theta$ that appear in $E_i(\xb)$. Let
\begin{equation}
    \langle \cdot \rangle = \frac{1}{n} \sum_{k=1}^{n} (\cdot)
\end{equation}
denote the empirical average over samples $\{\xb^{(k)}\}_{k=1,\ldots,n}$. The PL approach aims to maximize the logarithm of the conditional distribution \eqref{eq:Conditional_distribution} as defined next.

\begin{definition}[PL Estimator]
The Pseudo-Likelihood estimator is defined as a series of independent local optimization problems over $p$ nodes:
\begin{align}
    \widehat{\Theta}^{\text{PL}}_i &= \argmin_{\Theta_i} - \left\langle \log P(x_i \mid \xb_{\backslash i}) \right\rangle
    \\
    &= \argmin_{\Theta_i} \left\langle - E_i(\xb) + \log Z_i \right\rangle.
    \label{eq:Pseudo-Likelihood_Minimization}
\end{align}
\end{definition}

Notice that due to the local form of the estimator \eqref{eq:Pseudo-Likelihood_Minimization}, each component of the symmetric array $\Theta$ is estimated independently from each node's neighborhood. As is common in local node-wise regressions, a single final estimate of the parameter is obtained by a consensus average of independent local estimates.
In this work, we use the geometric mean to produce the final estimate of the components of $\Theta$, following the arguments in \cite{misra2020information} that showed that this choice of symmetrization is optimal for Gaussian graphical models. For instance, the final estimate for a pairwise coupling $\widehat{\theta}^{\text{PL}}_{ij}$ will be given by $\sqrt{\widehat{\theta}^{\text{PL}}_{ij} \widehat{\theta}^{\text{PL}}_{ji}}$, the geometric mean of couplings estimated from regressions based on the neighborhoods of $i$ and $j$.

The minimization in \eqref{eq:Pseudo-Likelihood_Minimization} can be done by gradient descent, where at each iteration one needs to re-evaluate integrals of the type \eqref{eq:Integral_Normalization_Conditional} by means of numerical integration, which can be costly if the integral needs to be computed to a high degree of precision. In the special case of Gaussian graphical models, $Z_i$ can be computed explicitly, $Z_i = \sqrt{\theta_{ii}/\pi}$, and hence numerical integration is not required. In this local form, the PL estimator is intimately connected to a node-wise regression approach introduced and analyzed for GGMs in \cite{meinshausen2006high}, which reduces to a simple least-squares regression problem upon a simple change of variables.

\subsection{Interaction Screening}

The Interaction Screening estimator has been first introduced for Ising models in \cite{Vuffray2016nips}, and recently generalized to the case of arbitrary discrete graphical models with multi-body interactions in \cite{vuffray2019efficient} where the Generalized Interaction Screening Objective (GISO) was introduced. In the case of general continuous distributions considered here, the Interaction Screening approach becomes particularly appealing due to its simple functional form that allows one to avoid the estimation of the local normalization factor $Z_i$. However, the GISO introduced in \cite{vuffray2019efficient} does not directly apply to distributions with unbounded support. This is because first, the so-called centered basis functions (see below) that are required to construct the GISO may not exist due to diverging integrals. Second, even when the GISO can be constructed, all its moments may not be finite leading to poor scaling of its sample complexity.

To alleviate these issues, we propose a modified estimator based on the GISO that is suitable for general distributions. The key idea is the introduction of a \emph{multiplicative regularizing distribution} (MRD) to make the GISO well behaved. We start by defining the estimator and discussing the intuition behind it, and then introduce a convenient choice of an MRD.

\subsubsection{Definition}

For each basis term $f_k(\xb_k)$ in the energy function \eqref{eq:General_Energy_Function} and each variable $x_i$, we define the \emph{local centering function} $g_{ik}(\xb_k)$, centered with respect to the MRD $R_i(x_i)$ as follows:
\begin{equation}
    g_{ik}(\xb_k) = f_k(\xb_k) - \int_{-\infty}^{+\infty} dx_i f_k(\xb_k) R_i(x_i),
    \label{eq:Cenetered_Factors}
\end{equation}
where we assume that the distribution $R_i(x_i)$ is chosen in such a way that the integral term in the expression \eqref{eq:Cenetered_Factors} exists. As suggested by its name, the local centering function sums to zero with respect to the MRD:
\begin{equation}
    \int dx_i g_{ik}(\xb_k) R_i(x_i) = 0.
    \label{eq:centering_property}
\end{equation}
Finally, for each node $i$ define the \emph{centered partial energy function} $E_i^g(\xb)$ as follows: 
\begin{equation}
    E_i^g(\xb) = - \sum_{k \in \mathcal{K}_i} \theta_k g_{ik}(\xb_k).
    \label{eq:Centered_Energy_Function}
\end{equation}

Now we are ready to formulate the GISO-based objective function, which we refer to as \emph{Interaction Screening Objective for Distributions with Unbounded Support} (ISODUS).

\begin{definition}[ISODUS Estimator]
The ISODUS estimator is defined as a series of independent local optimization problems over $p$ nodes:
\begin{equation}
    \widehat{\Theta}^{\text{IS}}_i = \argmin_{\Theta_i} \left\langle \exp\left(- E^g_i(\xb) \right) R_i(x_i) \right\rangle,
    \label{eq:Interaction_Screening_Objective}
\end{equation}
where $E^g_i(\xb)$ is given in \eqref{eq:Centered_Energy_Function}.
\end{definition}
Similarly to the post-processing step for the PL estimator, we use the geometric mean of model parameters estimated from different neighborhoods to produce the final estimate of $\Theta$ after the minimization has been performed.

ISODUS, as its name suggests, is constructed on the basis of the ``interaction screening'' principle \cite{Vuffray2016nips}. As a consequence of this property, in the limit of a large number of samples, the unique minimizer of the convex ISODUS objective is realized at the true value of the distribution parameters, regardless of a specific form of the MRD $R_i(x_i)$, as long as the estimator is correctly defined. A simple demonstration of this fact which exploits the property \eqref{eq:centering_property} is given in Appendix \ref{app:interaction_screening_property}.

The original GISO introduced in \cite{vuffray2019efficient} for discrete graphical models can be regarded as a special case of \eqref{eq:Interaction_Screening_Objective} where the MRD term $R_i(x_i)$  chosen as a uniform distribution over discrete alphabet of values that $x_i$ can take. It is immediately apparent that for the uniform MRD the integral in \eqref{eq:Cenetered_Factors} does not exist even in the principled value sense whenever $f_k$ contains an even degree monomial in $x_i$, an issue that arises even for the simple case of normal distributions.

\subsubsection{Specific form of the MRD}

We propose the following choice of MRD with unbounded support which decays fast enough such that the ISODUS estimator \eqref{eq:Interaction_Screening_Objective} is properly defined:
\begin{equation}
    R_i(x_i) = \frac{(s_i+\delta_i)}{2 \Gamma\left(\frac{1}{s_i+\delta_i}\right)}\nu_i^{\frac{1}{s_i+\delta_i}}  \exp\left(- \nu_i \vert x_i \vert^{s_i+\delta_i}\right),
    \label{eq:Prior}
\end{equation}
where $s_i$ is the maximum monomial order in the partial energy function \eqref{eq:Centered_Energy_Function}, $\nu_i, \delta_i > 0$ are the user-defined hyperparameters, $Z^i_p$ is the normalization constant, and $\Gamma(\cdot)$ is the Gamma function.

Observe that since the basis functions $f_k$ are polynomials, the centering functions in \eqref{eq:Cenetered_Factors} always exist, thus addressing the first issue with GISO. Also for any $\nu_i, \delta_i > 0$ the term $- \nu_i \vert x_i \vert^{s_i+\delta_i}$ dominates all other monomials in the GISO exponent, leading to all its expected moments being finite (see Appendix \ref{app:finiteness_moments} for a simple argument). For simplicity, in what follows we will choose $s_i = s$, where $s$ is the maximum monomial order in the full energy function \eqref{eq:General_Energy_Function}, and choose uniform values for the remaining hyper-parameters across all nodes, $\nu_i = \nu$ and $\delta_i = \delta$ for all $i \in V$.

The choice of MRD in \eqref{eq:Prior} is motivated by convenience and simplicity: it has the most natural and minimal exponential form for energy functions in monomial basis, where an addition of a term $-\vert x_i \vert^{s+\delta}$ makes ISODUS estimator well-defined. Moreover, for the exponential form of the distribution in Eq. \eqref{eq:Prior} and in the case of the energy function \eqref{eq:General_Energy_Function} expressed in the monomial basis, the centering functions in Eq.~\eqref{eq:Cenetered_Factors} can be computed analytically, which removes the need for the numerical integration. Since $R_i(x_i)$ is symmetric, all terms $f_k(\xb_k)$ where $x_i$ appears at the odd power are already centered, i.e. for them $g_{ik}(\xb_k) = f_k(\xb_k)$. So the only nontrivial terms are those that only contain $x_i^l$ with $l$ even (e.g. $l=2$ or $l=4$ in the example \eqref{eq:General_Energy_Function_Example} above with up to fourth order interactions). For those terms, $g_{ik}(\xb_k) = f_k(\xb_k) - c^{(l)}_i f_k(\xb_k) / x_i^l$, where the centering coefficients $c^{(l)}_i$ can be computed as follows:
\begin{equation}
    c^{(l)}_i \equiv \int dx_i x_i^l R_i(x_i)
    = \frac{\nu^{-\frac{l}{s+\delta}} \Gamma\left(\frac{l+1}{s+\delta}\right)}{ \Gamma\left(\frac{1}{s+\delta}\right)}.
\end{equation}
For instance, for the terms of the type $f_{iijm}(x_i,x_j,x_m) = x_i^2 x_j x_m$, we get $g_{iijm}(x_i,x_j,x_m) = x_i^2 x_j x_m - c^{(2)}_i x_j x_m$, and for $f_{iiii}(x_i) = x_i^4$, we obtain $g_{iiii}(x_i) = x_i^4 - c^{(4)}_i$.

In experiments for both PL and ISODUS, we use the optimization software Ipopt \cite{wachter2006implementation} with tolerance $10^{-8}$ within the Julia/JuMP modeling framework for mathematical optimization \cite{DunningHuchetteLubin2017}.\footnote{The code is available at \url{https://github.com/lanl-ansi/isodus/}.}

\section{Performance on Gaussian distributions}
While the main computational advantages of ISODUS compared to PL become prominent for models with higher order interactions, we first consider the simple case of Gaussian graphical models where generation of independent samples is easy for analyzing certain fundamental properties of ISODUS. We first show that when a sufficient regularization is imposed, ISODUS can successfully recover GGMs with appropriately decreasing error as more samples are available. We also show that the value of hyperparameters $\nu$ and $\gamma$ have little influence on the estimation accuracy (and no impact on the scaling) as long as a sufficient regularization is achieved. We then consider the so-called high dimensional setting for sparse GGMs and show that ISODUS with an $\ell_1$ sparsity regularization attains the well-known $\log p$ scaling in sample complexity.
 
\subsection{Reconstruction properties of ISODUS}

We begin by studying the influence of hyperparameters $\delta$ and $\nu$ of the MRD \eqref{eq:Prior} on the performance of the estimator. For the Gaussian energy function \eqref{eq:Gaussian_energy_function}, we have the maximum monomial order $s=2$. We quantify the parameter learning accuracy using the mean absolute reconstruction error $\epsilon \equiv \frac{1}{\vert \Theta \vert} \sum_k \vert \widehat{\theta}_k - \theta_k \vert$, where $\vert \Theta \vert$ is the number of components of $\Theta$. Here, we study a Gaussian distribution on 10 nodes, with a randomly-generated symmetric and positive semi-definite precision matrix.

\begin{figure}[!htb]
    \centering
    \includegraphics[width=0.5\columnwidth]{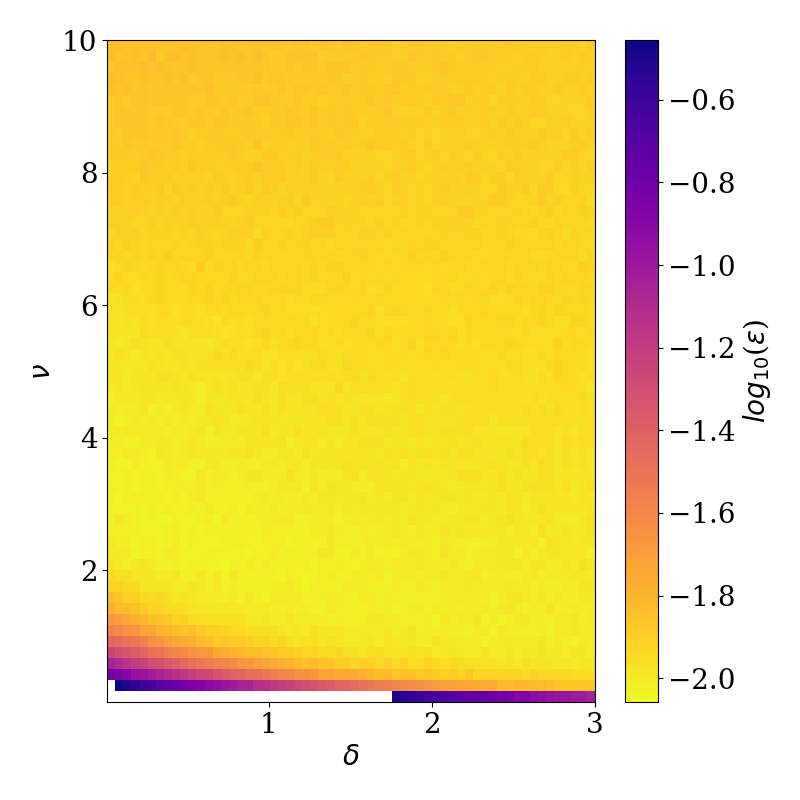}
     \caption{Mean absolute reconstuction error $\epsilon$ as a function of hyperparameters $\delta$ and $\nu$ for a GGM learning problem on 10 nodes. Each axis has 50 steps, and each point has been averaged over 80 independent sets of $10^5$ samples. The region in white is not shown as it results in high reconstruction errors.}
    \label{fig:sweep_nu_delta}
\end{figure}

In Figure \ref{fig:sweep_nu_delta}, we sweep over a range of $\delta$ and $\nu$ for a single instance of the GGM learning problem on 10 nodes and plot the resulting reconstruction error. From Figure \ref{fig:sweep_nu_delta}, it is apparent that in a region that is close to $\delta=0$, ISODUS is not regularized enough by the MRD and its reconstruction errors blow up dramatically.

In a sufficiently regularized region, the accuracy landscape looks like a valley with a weak dependence on both coefficients $\nu$ and $\delta$: as the MRD regularization becomes stronger with growing $\delta$, almost any coefficient $\nu > 0$ leads to a small reconstruction error. In what follows and for definiteness, we fix the value of these hyperparameters to values in the intermediate range, $\delta = 2$ and $\nu = 2$, which ensure that the MRD is not too strong, while the dependence on the optimal value of $\nu$ remains weak. As we will see later, with this arbitrary but fixed choice of hyperparameters, we recover all efficient sample scaling properties found for instance with the PL estimator. Since our main goal is studying the scaling properties of ISODUS, in what follows we don't optimize over the hyper-parameters even though it could have potentially reduced the absolute reconstruction error.

\begin{figure}[!htb]
    \centering
    \includegraphics[width=0.5\columnwidth]{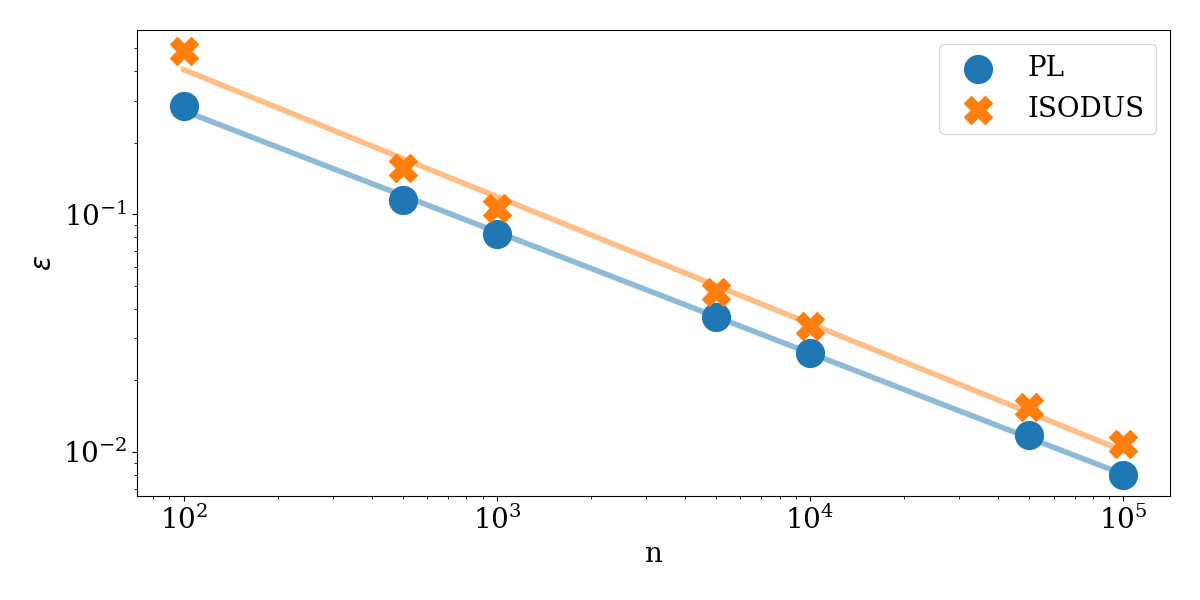}
    \caption{Scaling of the mean absolute error in the reconstruction of model parameters with number of samples for PL and ISODUS for a GGM learning problem on 10 nodes. Each data point has been averaged over 25 independent sets of samples.}
    \label{fig:gaussian_isodus_sample_scaling}
\end{figure}

For the chosen regularization values, we show in Figure \ref{fig:gaussian_isodus_sample_scaling} that ISODUS enjoys the expected asymptotic scaling where the learning error decays with the number of samples as $\epsilon \propto 1/\sqrt{n}$, and exhibits similar absolute accuracy compared to the error obtained by the PL estimator. Such a scaling provides an important empirical evidence for the consistency of the estimator without any additional unnecessary assumptions, which is the setting that has been the focus in recent work \cite{Vuffray2016nips,Klivans2017,vuffray2019efficient,shah2020learning}.

\subsection{Learning of the structure of sparse high-dimensional Gaussian distributions}

One important property of estimators for sparse graphical models in the high-dimensional regime consists in a weak dependence of the sample requirement on the dimension of the problem. This is characterized by the number of samples required for model selection $n^*$ scaling only logarithmically with the dimension of the problem: $n^* \propto \log p$. Such high-dimensional estimators typically enforce sparsity via sparsity-promoting regularization. The convexity-preserving $\ell_1$ regularization is a popular choice. In this section, we empirically verify that when supplemented with a sparsity regularizer, ISODUS exhibits a high-dimensional scaling $n^* \propto \log p$.

A typical setting in which such a scaling is conveniently derived is the structure learning problem, where the goal is to reconstruct the sparse support of the distribution. We adapt the same setting in this section.
Notice that the structure learning problem is tightly related to that of parameter estimation: parameters recovered to a sufficient accuracy can be thresholded at a fraction of the minimum coupling which would recover the structure of the graph. This is the approach that we use in what follows.

The minimum number of samples for which the hyper-graph can be recovered with high probability is denoted by $n^*$. We design the following numerical experiment for checking the scaling of $n^*$ as a function of problem size $p$. We consider a family of Gaussian graphical models of different sizes $p$ supported on random regular graphs of degree $d=3$. The precision matrix $\Theta$ has ones on the diagonal, and the strength of each non-zero edge is equal to $\kappa = 0.25$, which ensures that the resulting $\Theta$ is positive semi-definite. We generate i.i.d. samples for each family of distributions, and perform reconstructions of $\Theta$ using both estimators, PL and ISODUS, supplemented with sparsity-promoting $\ell_1$ regularizers. With this modification, the Regularized ISODUS takes the following form:
\begin{align}
        \widehat{\Theta}^{\text{IS}}_i = \argmin_{\Theta_i} \Big(
        &\left\langle \exp\left(- E^g_i(X) - \nu \vert x_i \vert^{s+\delta}\right) \right\rangle 
        \\
        & + \lambda_{\text{IS}} \sqrt{\frac{\log{p}}{n}} \sum_{k \in \mathcal{K}_i} \vert \theta_{k} \vert \Big),
        \label{eq:regularized_isodus}
\end{align}
and the Regularized PL estimator has a similar form with its respective regularization coefficient $\lambda_{\text{PL}}$. In order to fix the regularization coefficients $\lambda_{\text{IS}}$ and $\lambda_{\text{PL}}$, we first run a preliminary study for $p=50$ and sweep over the sparsity-promoting hyperparameters to determine the optimal values for each estimator. Reconstruction error as a function of regularizer values is presented in Appendix \ref{app:structure_learning}. For the remainder of the experiment, we fix the values of regularization coefficients to optimal values $\lambda_{\text{IS}} = 0.35$ and $\lambda_{\text{PL}} = 2.3$ obtained in this experiment. Similarly to the selection of $\nu$ and $\delta$ in MRD, a non-optimal choice of $\lambda$ would preserve all desirable scalings, potentially at the expense of a constant overhead in the sample complexity, as we discuss in Appendix \ref{app:structure_learning}.

Following this, we use regularized versions of PL and ISODUS to determine $n^{*}$ for each $p$ using the fixed values of sparsity-promoting hyperparameters. For each $p$, $n^{*}$ is determined as follows: For a given number of samples $n$, once the symmetrized $\Theta$ is estimated, we threshold inferred pairwise coupling values at $\kappa/2 = 0.125$, i.e. only keeping $\theta_{ij} > \kappa/2$. The non-zero elements $\theta_{ij}$ in the precision matrix then define the reconstructed adjacency matrix of the graph, which can be compared to the underlying graph of the model. Let us define $n^{*}$ as the smallest number of samples such that the graph is reconstructed correctly for $45$ different and independent sets of samples in a row, which guarantees graph recovery with probability greater than $0.95$ with at a statistical confidence of at least $90\%$  \cite{lokhov2018science}. We use sequential search to determine $n^{*}$, increasing the search value by 25 samples if 45 trials in a row are not successful, and decreasing by 10 if they are.

\begin{figure}[!htb]
    \centering
    \includegraphics[width=0.5\columnwidth]{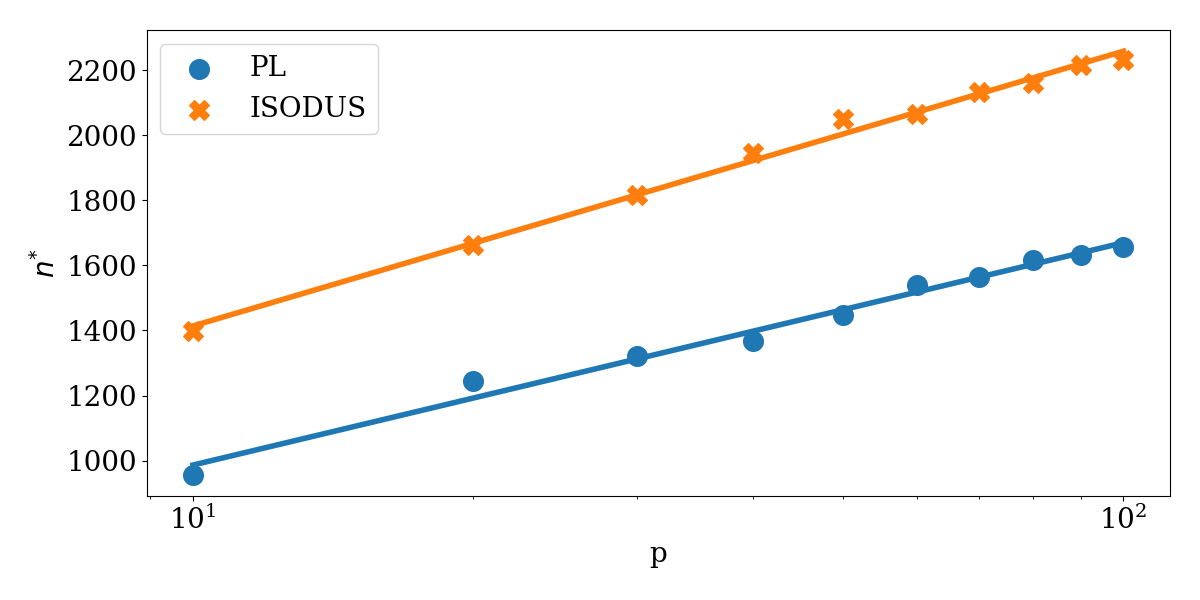}
    \caption{Scaling of required number of samples $n^{*}$ for structure learning problem in GGMs on random regular graphs with degree $d=3$  as a function of problem size $p$ using the optimal values of the sparsity-promoting regularization coefficients. Each data point is averaged over $5$ independent graph realizations of size $p$. For each model, $n^{*}$ is determined as the minimum number of samples that leads to a successful graph recovery for 45 independent sets of samples. Both regularized PL and ISODUS estimators show logarithmic scaling $n^{*} \propto \log p$, desirable for the high-dimensional structure learning problem.}
    \label{fig:log_scaling}
\end{figure}

The resulting scaling of $n^*(p)$ is presented in Figure \ref{fig:log_scaling}, where we further average over $5$ independent graph realizations for each $p$. We see that both regularized PL and ISODUS show logarithmic scaling $n^* \propto \log p$ with the dimension. All existing theoretical analysis establishing the $\log p$ scaling heavily rely on Chernoff type concentration bounds that require all moments of the estimator to be finite. Our choice of MRD in \eqref{eq:Prior} ensures finiteness of all moments and the empirical $\log p$ scaling observed in this section supports this choice.

\vspace{-0.2cm}

In Appendix \ref{app:structure_learning}, we also verify that the structure learning through tresholding of consistently learned parameters also holds for graphical models which do not satisfy irrepresentability condition, and for which sparsistency recovery fails. This provides an empirical evidence that ISODUS achieves consistent structure and parameter recovery without unnecessary assumptions such as incoherence or irrepresentability conditions.

\section{Tests on general non-Gaussian exponential family distributions}

In this section, we illustrate the advantages of ISODUS for learning general non-Gaussian distributions compared to PL. First, we consider a one-dimensional exponential family distribution with higher-order interactions, and compare ISODUS and PL both in terms of their accuracy, as well as in computational complexity. This family represents the simplest non-Gaussian case where we have to resort to numerical integration for computing the normalization factor $Z_i$ that enters in the PL estimator. Finally, we illustrate that ISODUS retains its desirable computational and sample scaling properties on more complex multivariate distributions with multi-body interactions.

\subsection{One-dimensional non-Gaussian distribution}

As an example of a one-dimensional non-Gaussian distribution, we take an exponential family \eqref{eq:General_Distribution} with the energy function $E(X) = -x^{2} - 0.5 x^3 - 2 x^4$. Unlike for GGMs, sample generation for models with higher-order moments is non-trivial. To ensure high-quality independent and identically distributed samples, we employ brute-force sampling using fine discretization with 5000 bins.

\begin{figure}[!htb]
    \centering
    \includegraphics[width=0.5\columnwidth]{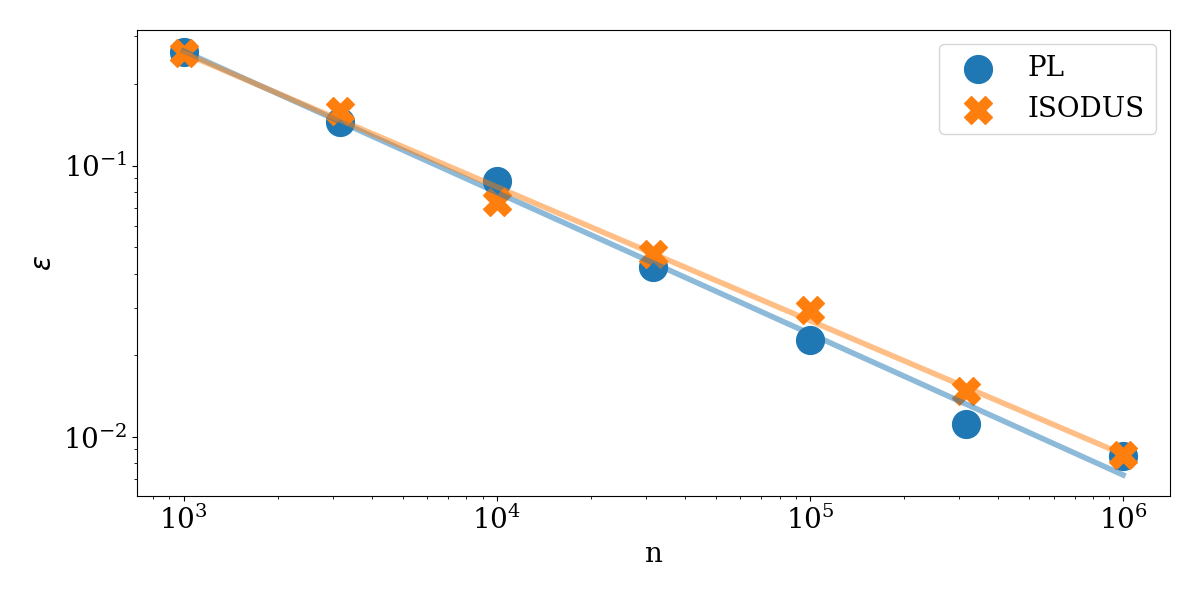}
    \caption{Scaling of the mean absolute error in reconstruction of model parameters with the number of samples for PL and ISODUS for a test one-dimensional distribution with the energy function of degree 4. Each point has been averaged over 45 independent sets of samples.}
    \label{fig:1D_error_scaling}
\end{figure}

As a first experiment, we compare the accuracies of ISODUS and PL for this test case. The results on the learning error are presented in Figure \ref{fig:1D_error_scaling}.
We used a relative tolerance of factor $10^{-7}$ in the evaluation of the normalization factor $Z$ at each step of the optimization procedure. Both algorithms demonstrate similar accuracy, and asymptotically normal error decay with the number of samples.

\begin{figure}[!htb]
    \centering
    \includegraphics[width=0.5\columnwidth]{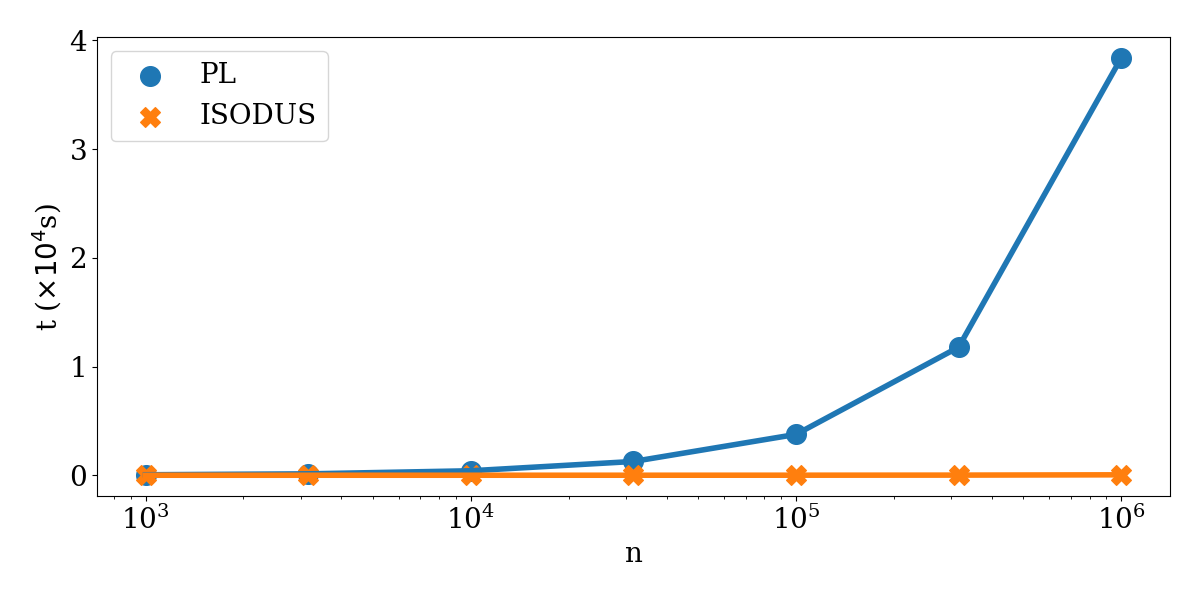}    \caption{Comparison of run-times of PL and ISODUS estimators in the case of one-dimensional non-Gaussian distribution as a function of the number of samples. PL run-time includes a computational overhead due to the necessity of numerical estimation of local normalization factors.
    }
    \label{fig:1D_computational_complexity}
\end{figure}

\begin{figure}[!htb]
    \centering
    \includegraphics[width=0.5\columnwidth]{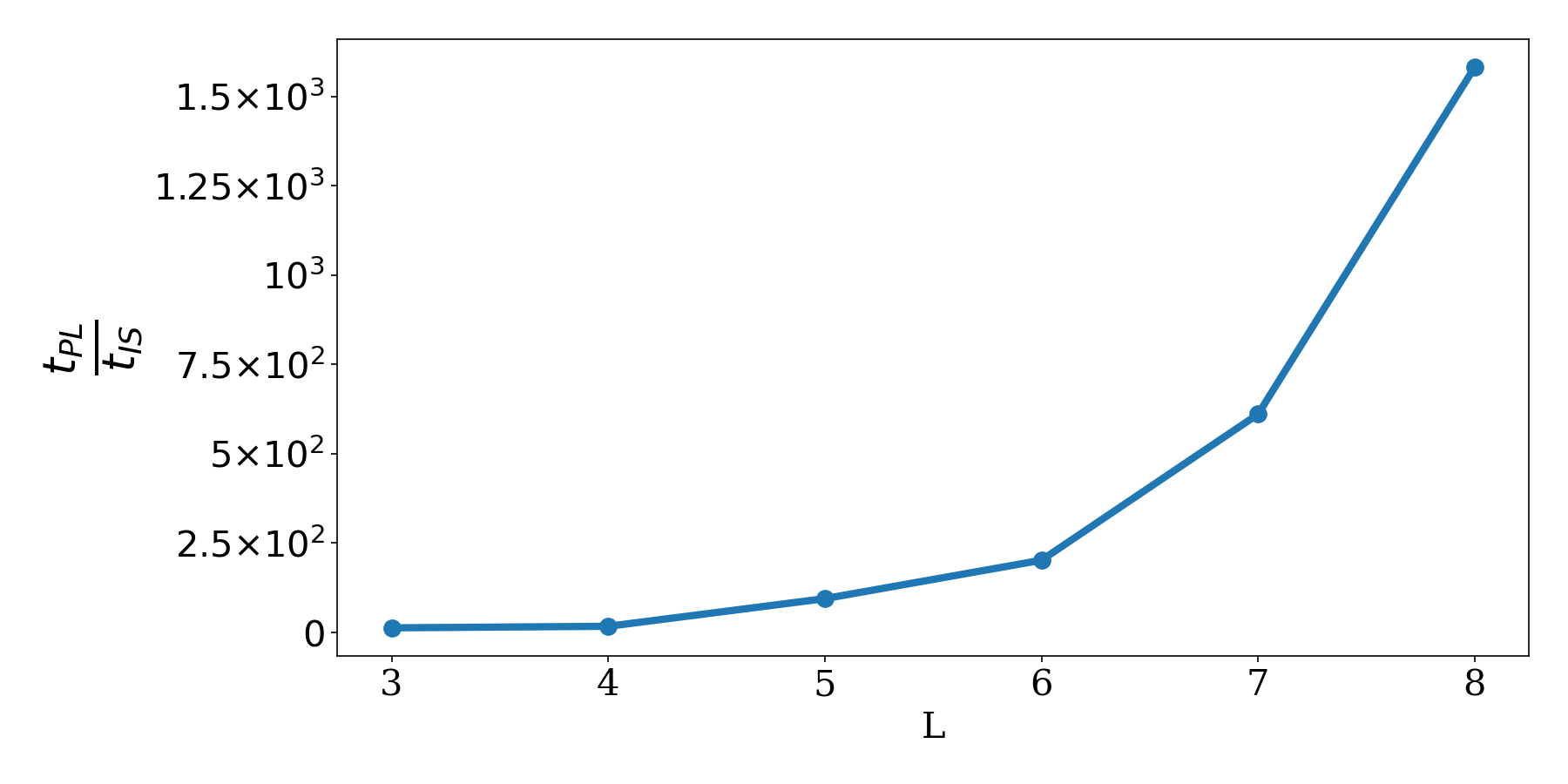}
    \caption{Dependence of the ratio of running times of PL and ISODUS as a function of the maximum degree of the polynomial energy function $L$ for one-dimensional non-Gaussian distributions.}
    \label{fig:1D_energy_function_complexity}
\end{figure}

As the next test, we check the pure computational run-time of both algorithms on this instance as a function of the number of samples. The comparison of run-times is presented in Figure \ref{fig:1D_computational_complexity}. The difference in running time is explained by the additional overhead of numerical estimation of the local normalization factors required during the PL optimization subroutine. This numerical estimation up to a given precision has been performed using the QuadGK.jl library based on the adaptive Gauss-Kronrod quadrature \cite{QuadGK}. It is natural to assume that the complexity of adaptive numerical integration up to a prescribed tolerance may increase with the complexity of the integrand. To test this hypothesis, we ran a comparison of run-times of both algorithms for learning one-dimensional distributions with energy functions of increasing monomial order $L$. The ratio of run-times of PL and ISODUS for a fixed number of samples $n=100$ is plotted in Figure \ref{fig:1D_energy_function_complexity} for the models with the energy function $E(X) = - \sum_{l=2}^L \alpha_{l} x^{l}$ where $\alpha_{l}$ have been randomly selected from $[0,1]$. We see that the computational overhead of PL may significantly increase with the complexity of the model.


\subsection{Multivariate higer-order distributions}

Results on one-dimensional distributions demonstrate that ISODUS shows similar reconstruction results in terms of accuracy when compared to the baseline PL method, but benefits from a much lower computational complexity. In fact, high run-time of PL makes comparisons for models beyond 2D distributions prohibitive. The most non-trivial model in which we could still carry an additional comparison of PL and ISODUS was a 2D non-Gaussian distributions with a general energy function with interactions order up to 4, which are important in some applications such as modeling scalar functions in two-dimensional turbulence. We report the results in Appendix \ref{app:multivariate_multibody} which look very similar to the ones for the 1D distributions: For a similar accuracy obtained for $10^5$ samples, ISODUS run-time is 9 seconds, and PL is 7,000 times slower.

Here, we illustrate the scalings of ISODUS on a four-dimensional problem with an energy function that contains higher-order interactions up to order 4. Production of independent samples is the main bottleneck for running tests in higher-dimensional problems. To ensure the quality of the samples remains high, we aim to maintain the same discretization level (5000 points per dimension) in this problem as well, which leads to an intractable resolution for general four-dimensional distributions. We chose to construct a pseudo four-dimensional distribution, which factorizes over two pairs of variables. Hence, i.i.d. samples from the ground-truth distribution could be constructed from a combination of two independent 4-body distributions over two variables each. The ISODUS estimator was not fed with this information, and had to learn the structure of the model from the provided four-dimensional samples. The scaling of the reconstruction error as a function of number of samples is given in the Figure \ref{fig:4D_error_scaling}. PL's run-time renders the algorithm essentially intractable for this problem. In Appendix \ref{app:multivariate_multibody} we also verify that, as expected, the run-time of ISODUS is linear as a function of number of samples.

\begin{figure}[!htb]
    \centering
    \includegraphics[width=0.5\columnwidth]{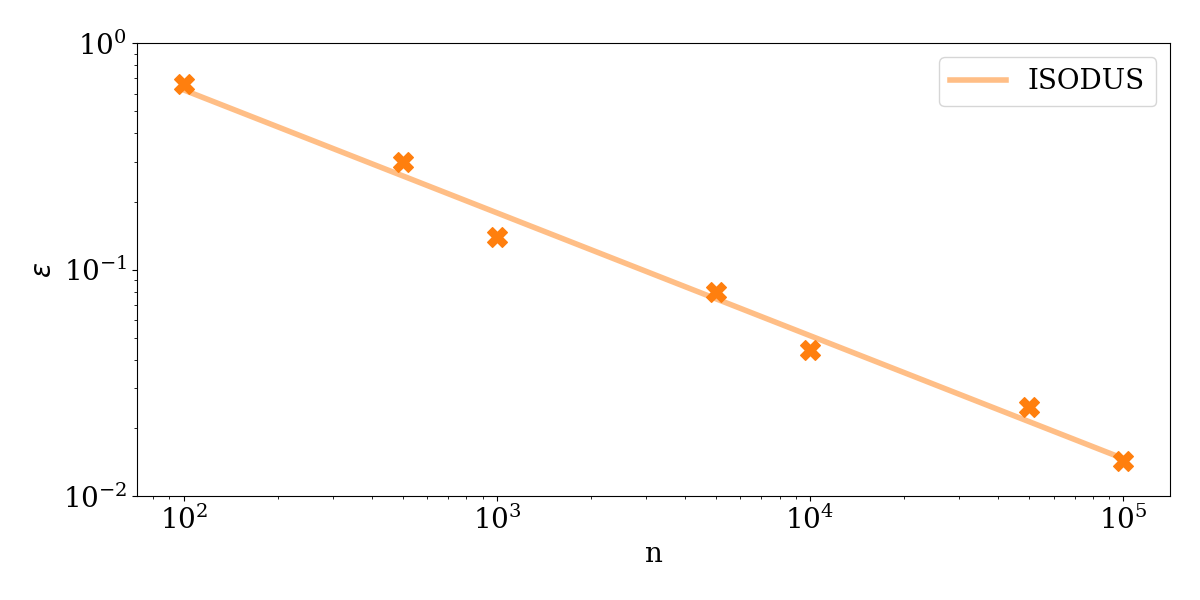}
    \caption{Scaling of the reconstruction error of ISODUS as a function of number of samples on a four-dimensional four-body distribution. ISODUS shows an expected scaling $\epsilon \propto 1/\sqrt{n}$. Each point has been averaged over 45 independent sets of samples.}
    \label{fig:4D_error_scaling}
\end{figure}

\section{Conclusion}

In this paper, we introduced a new estimator, ISODUS, adapted from the Interaction Screening method to the challenging case of general exponential family distributions with unbounded support. We studied the properties of this estimator for Gaussians as well as for higher-order distributions, and established empirically that it shows an error dependence with respect to the number of samples similar to the Pseudo-Likelihood approach. However, ISODUS avoids computing local normalization factors that makes the PL algorithm computationally prohibitive. This makes ISODUS the estimator of choice for learning continuous non-Gaussian graphical models in practice.

We have shown how the inclusion of sparsity-promoting regularization allows one to recover the high-dimensional scaling $n^{*} \propto \log p$ for structure and parameter estimation problems.
We have focused on distributions with higher-order terms in monomial form which are important in applications where non-Gaussian statistics are relevant. It would be interesting to study ISODUS for other families of sufficient statistics, or even for implicit parametrizations of the energy function such as neural networks introduced recently for discrete graphical models 
\cite{abhijith2020learning}.

\section*{Acknowledgements}
We acknowledge support from the Laboratory Directed Research and Development program of Los Alamos National Laboratory under project numbers 20190059DR, 20200121ER, 20210078DR, and 20210674ECR.

\vskip 0.2in
\bibliography{biblio}

\appendix

\renewcommand{\theequation}{S\arabic{equation}}

\setcounter{equation}{0}

\renewcommand{\thefigure}{S\arabic{figure}}

\setcounter{figure}{0}

\renewcommand\appendixname{}

\section{Interaction Screening property for ISODUS}
\label{app:interaction_screening_property}

Here, we present a simple derivation which shows that in the limit of large number of samples, the unique minimizer of the convex ISODUS objective is achieved at $\widehat{\Theta} = \Theta$, where the true local interactions present in the model are fully ``screened'' Moreover, this happens irrespectively of a concrete form of the MRD $R_i(x_i)$, as long as the estimator is correctly defined. For the particular choice in \eqref{eq:Prior}, this result does not depend on particular values of $\nu$ and $\delta$ as long as they are positive. To this end, we will show that components of the gradient of the ISODUS objective is zero at the value of the true parameters of the model $\Theta$.

In the limit $n \to \infty$, the empirical average in \eqref{eq:Interaction_Screening_Objective} becomes an average with respect to the measure \eqref{eq:General_Distribution}. Choose a $k \in \mathcal{K}_i$ and select the respective $\theta_k$. Then the gradient component with respect to $\theta_k$ reads
\begin{align}
    \frac{\partial}{\partial \theta_k} & \lim_{n \to \infty} \left\langle \exp\left(- E^g_i(\xb) \right) R_i(x_i) \right\rangle
    \\
    & = \int d\xb \mu(\xb) g_{ik}(\xb_k) \exp\left(- E^g_i(\xb) \right) R_i(x_i)
    \\
    & = \int d\xb_{\backslash i} \int dx_i g_{ik}(\xb_k) \underbrace{\exp\left(- \sum_{k \in \mathcal{K}} \theta_k f_k(\xb_k) \right) \exp\left( \sum_{k \in \mathcal{K}_i} \theta_k g_{ik}(\xb_k) \right)}_{\text{independent of $x_i$}} R_i(x_i) = 0
\end{align}
because the local centering functions \eqref{eq:Cenetered_Factors} sum to zero with respect to MRD as in \eqref{eq:centering_property}.

\section{Finiteness of expected moments of ISODUS}
\label{app:finiteness_moments}

Here, we use a minimal example of a two-dimensional Gaussian distribution to provide a simple argument which shows that the expected moments of ISODUS are finite for positive $\nu$ and $\delta$ in the MRD parametrization \eqref{eq:Prior}, while $\delta = 0$ provides an insufficient regularization.

Consider the distribution of the form
\begin{equation}
    \mu(\xb) \propto \exp(-x^2 -\theta^{*} xy -y^2).
\end{equation}
Let us consider local reconstruction for the node $x$. The expectation of the $m$th moment of ISODUS then reads
\begin{equation}
    \E[\left(\exp\left(- E^g_i(\xb) \right) R_i(x_i)\right)^m] \propto \int dx dy \exp(-x^2 -\theta^{*} xy -y^2) \exp\left( m (x^2 + \theta xy) + m \nu \vert x \vert^{2+\delta}  \right).
\end{equation}
Let us consider the case $\delta = 0$ first. In this case, the quadratic form in the exponential has the matrix form
\begin{equation}
    \begin{bmatrix}
    m\nu + m - 1 & \frac{m\theta - \theta^*}{2} \\
    \frac{m\theta - \theta^*}{2} & 1 
    \end{bmatrix}
    \label{eq:quadratic_form}
\end{equation}
and has a negative determinant for a large enough $m$. Hence for a ``minimal'' regularization $\delta = 0$ and any finite $\nu$, some higher-order moments of the estimator may not exist, which may potentially lead to poor sample complexity properties. This is however not the case for $\delta > 0$. Indeed, for any $m$, there exists a finite radius $r$ for which $\vert x \vert^{2+\delta} \geq x^2 r^\delta$ with the first diagonal element of the quadratic form \eqref{eq:quadratic_form} becoming $m \nu r^\delta + m -1$, which could be made sufficiently large to keep the matrix positive semidefinite, and hence keeping any moment finite.

\section{Additional considerations for structure learning}
\label{app:structure_learning}

\begin{figure}[!htb]
    \centering
    \includegraphics[width=0.5\columnwidth]{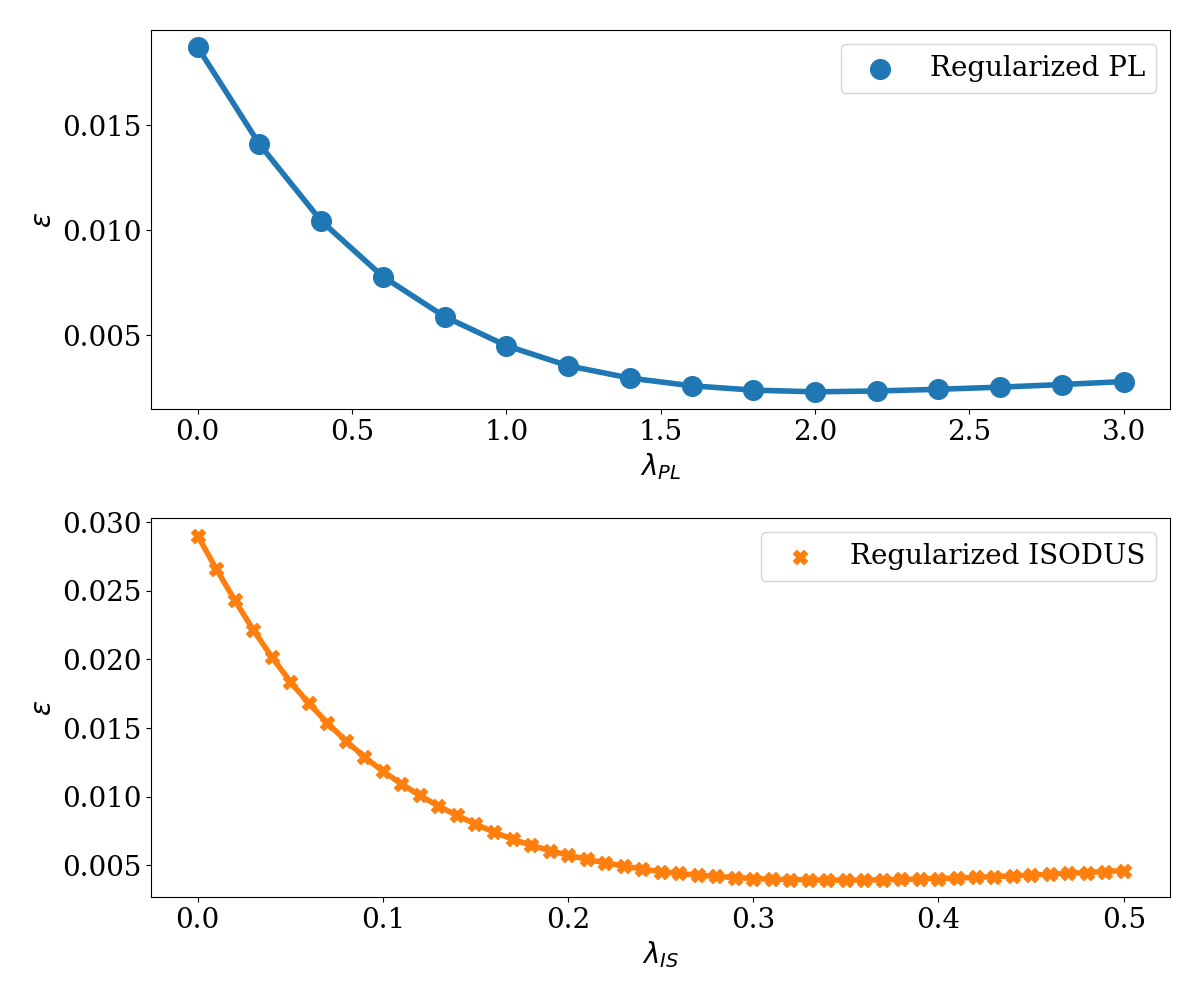}
    \caption{Sweep over $\lambda_{\text{IS}}$ and $\lambda_{\text{PL}}$ hyperparameters for regularized versions of PL and ISODUS for the GGM with $p=50$ nodes. The minima of the mean absolute reconstruction errors are obtained for $\lambda_{\text{IS}} = 0.35$ and $\lambda_{\text{PL}} = 2.3$.}
    \label{fig:lambda_sweep}
\end{figure}

Here, we show results of a sweep over sparsity hyper-parameters $\lambda_{\text{IS}}$ and $\lambda_{\text{PL}}$ in structure learning problem to determine the optimal values for each estimator. Reconstruction error as a function of regularizer values is presented in Figure \ref{fig:lambda_sweep}. The optimal values for $p=50$ are $\lambda_{\text{IS}} = 0.35$ and $\lambda_{\text{PL}} = 2.3$.

\begin{figure}[!h]
    \centering
    \includegraphics[width=0.5\columnwidth]{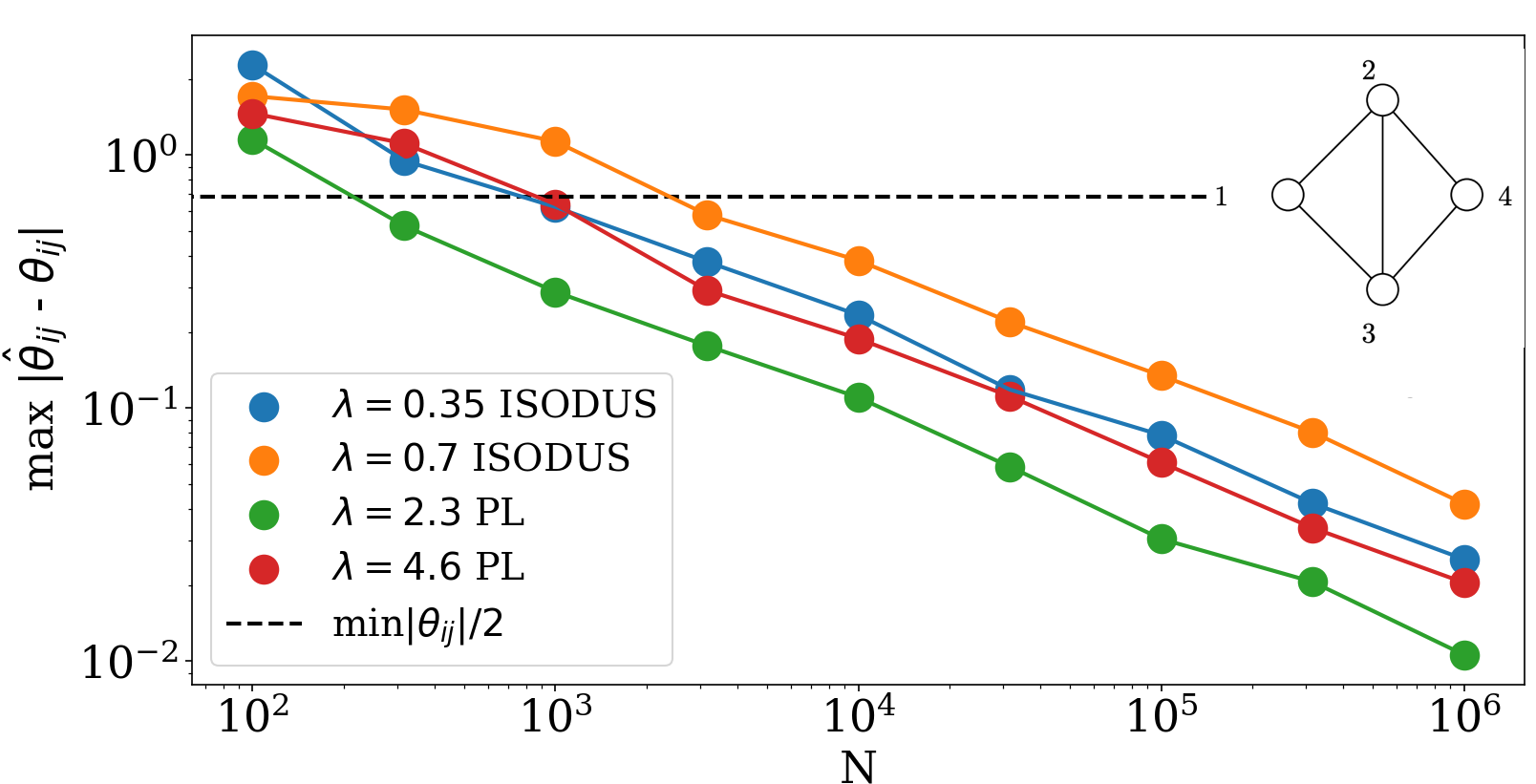}
    \caption{Example of a toy Gaussian model in the regime where \emph{sparistency} recovery provably fails. However both PL and ISODUS are \emph{consistent} for parameter learning, and the structure is recoverable once the max-error goes below the dashed line.}
    \vspace{-0.39cm}
    \label{fig:incoherence}
\end{figure}

Generally, the choice of hyper-parameters in graphical model selection is a common question. Most theorems establishing sample complexity bounds with regularization, e.g. \cite{ravikumar2010high,Vuffray2016nips,Klivans2017,vuffray2019efficient}, typically require $\lambda > \lambda^*$, where $\lambda^*$ is an explicit function of model parameters. Practically, this implies that it is more beneficial to ``over-regularize'' than ``under-regularize'', which would preserve all desirable scalings, potentially at the expense of a constant overhead in the sample complexity. We observe the same phenomenon in Figures \ref{fig:sweep_nu_delta} and \ref{fig:lambda_sweep}, where the estimation error becomes flat once sufficient regularization with $\nu(\delta)$ and $\lambda$ are reached. A non-optimal choice of the hyper-parameters does not change the scaling of the error rate. We additionally illustrate this point in Figure \ref{fig:incoherence} where the sparsity regularizer $\lambda$ is doubled: the scaling of the max error in the reconstruction decays with the same scaling rate at the expense of an extra constant factor of samples.

All recent works focus on a setting of \emph{consistency}, where learned parameters thresholded at a fraction of the minimum coupling recovers the structure of any model without unnecessary assumptions such as incoherence or irrepresentability, see e.g. \cite{vuffray2019efficient} for discrete or \cite{misra2020information} for Gaussian distributions. This is to differentiate with previous results in GGMs which studied the \emph{sparsistency} property: structural learning without thresholding of the parameters, which require additional assumptions on the graphical model. To illustrate this point, we used the diamond-structured graph example from \cite{ravikumar2008model}, and ran PL and ISODUS for $\rho=0.55$, i.e. in the regime where \emph{sparsistency} provably fails for both ML and LASSO. Figure \ref{fig:incoherence} shows that both PL and ISODUS are consistent.

\section{Learning of multivariate multi-body distributions}
\label{app:multivariate_multibody}

\begin{figure}[!htb]
    \centering
    \includegraphics[width=0.93\columnwidth]{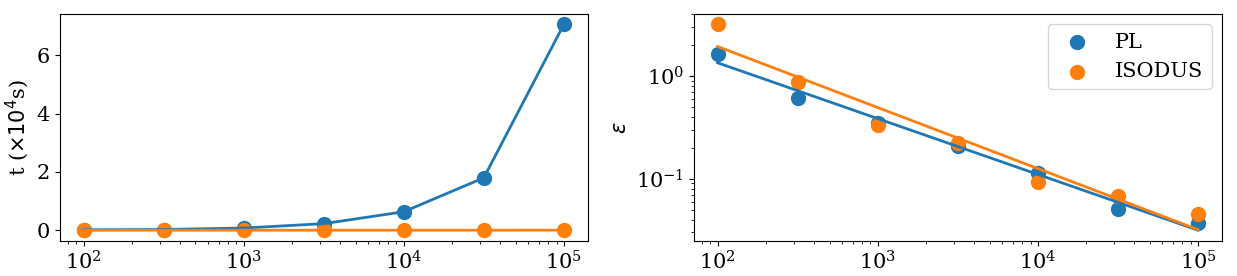}
    \caption{Comparison of the run-time (left) and accuracy (right) of PL and ISODUS for a non-Gaussian two-dimensional distribution with the energy function of degree 4. For a similar accuracy obtained for $10^5$ samples, ISODUS run-time is 9 seconds, and PL is 7,000 times slower. Each point has been averaged over 45 independent sets of samples.}
    \label{fig:2D_error_scaling}
\end{figure}

In Figure \ref{fig:2D_error_scaling}, we present results of comparison for the most complex model where PL is still tractable: a 2D non-Gaussian distribution with a general energy function with interactions order up to 4. While ISODUS's run time still scales linearly with the number of samples as it should be, the PL run-time is dominated by the time required to numerically evaluate the local partition function.

\begin{figure}[!htb]
    \centering
    \includegraphics[width=0.48\columnwidth]{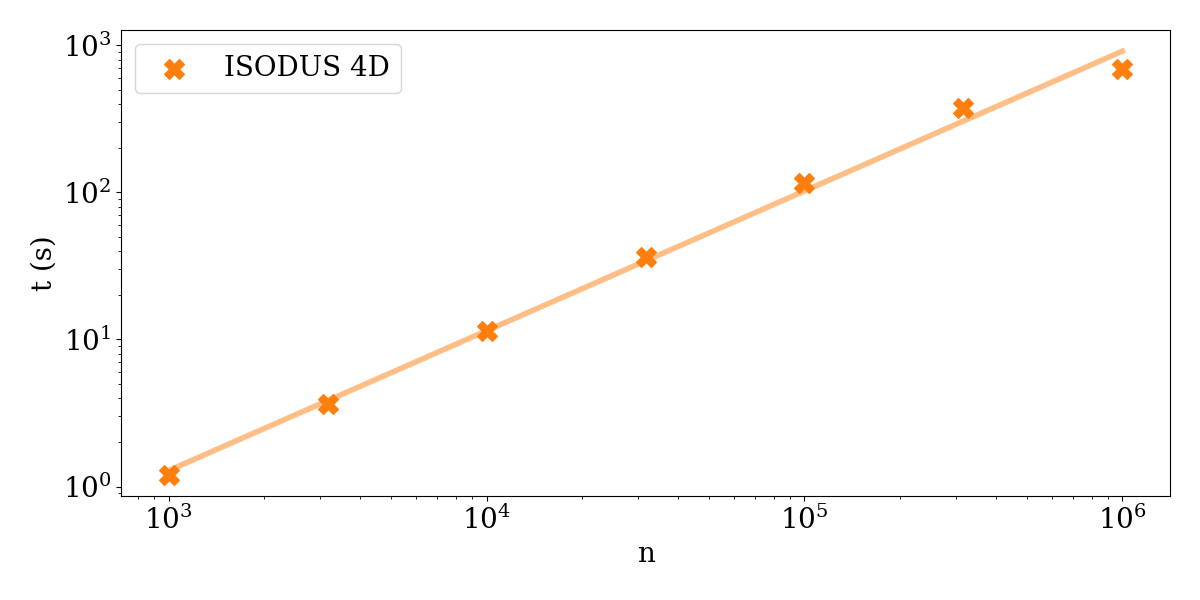}
    \caption{Linear run-time scaling of ISODUS with the number of samples for a four-dimensional four-body distribution. Each point has been averaged over 45 independent sets of samples.}
    \label{fig:4D_computational_complexity}
\end{figure}

The linear run-time scaling of ISODUS with the number of samples is further verified for a 4D distribution and fourth-order energy function in Figure \ref{fig:4D_computational_complexity}.

Details on the test-cases and code can be found at \url{https://github.com/lanl-ansi/isodus/}.

\end{document}